\documentclass[letterpaper]{article} % DO NOT CHANGE THIS
\usepackage[draft]{aaai2026}  % DO NOT CHANGE THIS
\usepackage{times}  % DO NOT CHANGE THIS
\usepackage{helvet}  % DO NOT CHANGE THIS
\usepackage{courier}  % DO NOT CHANGE THIS
\usepackage[hyphens]{url}  % DO NOT CHANGE THIS
\usepackage{graphicx} % DO NOT CHANGE THIS
\usepackage{natbib}  % DO NOT CHANGE THIS AND DO NOT ADD ANY OPTIONS TO IT
\usepackage{caption} % DO NOT CHANGE THIS AND DO NOT ADD ANY OPTIONS TO IT
\usepackage{algorithm}
\usepackage{algorithmic}

\usepackage{bm}
\usepackage{amsfonts}
\usepackage{amsmath}
\usepackage{booktabs}

\newtheorem{lemma}{Lemma}

\newtheorem{assumption}{Assumption}

\usepackage{newfloat}
\usepackage{listings}
\DeclareCaptionStyle{ruled}{labelfont=normalfont,labelsep=colon,strut=off} % DO NOT CHANGE THIS
\floatstyle{ruled}
\newfloat{listing}{tb}{lst}{}
\floatname{listing}{Listing}
\title{Optimal Growth Schedules for Batch Size and Learning Rate in SGD \\
that Reduce SFO Complexity}
\author{
 Hikaru Umeda,
 Hideaki Iiduka
}
\affiliations {
Meiji University \\
ee227115@meiji.ac.jp, iiduka@cs.meiji.ac.jp
}

\usepackage{bibentry}
\begin{document}

\maketitle

\begin{abstract}
The unprecedented growth of deep learning models has enabled remarkable advances but introduced substantial computational bottlenecks.
A key factor contributing to training efficiency is batch-size and learning-rate scheduling in stochastic gradient methods.
However, naive scheduling of these hyperparameters can degrade optimization efficiency and compromise generalization.
Motivated by recent theoretical insights, we investigated how the batch size and learning rate should be increased during training to balance efficiency and convergence.
We analyzed this problem on the basis of stochastic first-order oracle (SFO) complexity, defined as the expected number of gradient evaluations needed to reach an $\epsilon$–approximate stationary point of the empirical loss.
We theoretically derived optimal growth schedules for the batch size and learning rate that reduce SFO complexity and validated them through extensive experiments.
Our results offer both theoretical insights and practical guidelines for scalable and efficient large-batch training in deep learning.
\end{abstract}

\begin{links}
    \link{Code}{https://anonymous.4open.science/r/optimal-schedule}
\end{links}

\section{Introduction}

The rapid expansion of deep learning models has enabled substantial advances across a wide range of tasks, but this progress has come with increasing computational demands.
Achieving high performance across diverse tasks requires an large number of gradient evaluations and substantial computational resources, which renders training efficiency a key bottleneck in deep learning.
To address this problem, researchers have proposed such approaches as model pruning \cite{NIPS2015_ae0eb3ee, li2017pruning} and parameter-efficient fine-tuning \cite{pmlr-v97-houlsby19a}. Even with these approaches, however, large-scale training remains computationally expensive and resource intensive.

A key determinant of training efficiency in stochastic gradient methods is the joint setting of batch size and learning rate. Mini-batch stochastic gradient descent (SGD) \citep{robb1951,zinkevich2003,nem2009,gha2012,gha2013} and its variants remain the backbone of large-scale optimization due to their simplicity, scalability, and widespread applicability.
Using larger batches can exploit GPU parallelism more effectively and improve throughput. However, naively increasing the batch size often degrades the model’s generalization performance, leading to lower test accuracy—a phenomenon known as the \emph{generalization gap} \cite{keskar2017on}.
To address this, recent approaches use a dynamic scheduling strategy: begin training with a small batch size and gradually increase it over time \cite{Byrd:2012aa,balles2016coupling,pmlr-v54-de17a, l.2018dont, goyal2018accuratelargeminibatchsgd}.
This approach has demonstrated empirical advantages, and recent theoretical studies further suggest that jointly increasing the batch size and learning rate can improve the convergence rate of mini-batch SGD \cite{umeda2025increasing}.

Motivated by these insights, we investigated how the batch size and learning rate should be increased to achieve more efficient training while maintaining desirable convergence properties. 
In particular, we analyzed this problem through the lens of \emph{stochastic first-order oracle (SFO) complexity}, which quantifies the total number of gradient evaluations required to reach an $\epsilon$--approximate stationary point \citep{doi:10.1137/120880811, Ghadimi:2016aa, Imaizumi19062024}. 
This metric provides a principled way to measure the computational effort of stochastic optimization methods, making it well-suited for studying the trade-offs arising from dynamic hyperparameter schedules.

We theoretically characterized optimal growth schedules for the batch size and learning rate, elucidating how their joint increase affects both convergence efficiency and computational cost. 
To bridge theory and practice, we validated our insights through empirical experiments on standard deep learning benchmarks, confirming that the proposed schedules enhance training efficiency without compromising model accuracy.
Beyond advancing the theoretical understanding of dynamic hyperparameter schedules, our findings offer practitioners clear and effective strategies for scaling deep learning models.

\subsection{Contributions}

This work advances both the theoretical and practical understanding of how the batch size and learning rate should be scheduled during training to enhance the training efficiency of mini-batch SGD. The main contributions are as follows:

\begin{itemize}
    \item \textbf{Theoretical analysis of SFO complexity.} 
    We analyzed mini-batch SGD under standard smoothness and bounded-variance assumptions and explicitly characterized how the batch size and learning rate jointly affect the SFO complexity required to reach an $\epsilon$–approximate stationary point.

    \item \textbf{Optimal growth schedules for batch size and learning rate.} 
    We derived convergence bounds for various increasing schedules and identified the \emph{critical batch size} that minimizes SFO complexity. In particular, for an exponentially increasing schedule,
\[
b_m = b_0 \cdot \delta^m, \quad \eta_m = \eta_0 \cdot \gamma^m,
\]
we show that the optimal condition is approximately $\gamma^2 \approx \delta$, indicating that the batch size must scale with the square of the learning rate in order to achieve optimal efficiency.

    \item \textbf{From theory to practice: empirical validation.} 
    We translated the theoretical insights into practical schedules, including linear and exponentially increasing schedules for the batch size and learning rate, and validated them on standard deep learning benchmarks, including ResNet‑18 on the CIFAR‑100 dataset. The results demonstrate improved training efficiency and provide actionable guidelines for large‑batch training.
\end{itemize}

\section{Theoretical Background}

\subsection{Empirical Risk Minimization}

Let $\bm{\theta} \in \mathbb{R}^d$ denote the parameters of a deep neural network; let $S = \{(\bm{x}_1,\bm{y}_1), \ldots, (\bm{x}_n,\bm{y}_n)\}$ be the training set, where data point $\bm{x}_i$ is paired with label $\bm{y}_i$; and let $f_i (\cdot) := f(\cdot;(\bm{x}_i,\bm{y}_i)) \colon \mathbb{R}^d \to \mathbb{R}_+$ be the loss function for the $i$-th training example $(\bm{x}_i,\bm{y}_i)$. Empirical risk minimization minimizes the empirical loss defined for all $\bm{\theta} \in \mathbb{R}^d$ as
\begin{align*}
    f (\bm{\theta}) 
    = \frac{1}{n} \sum_{i\in [n]} f(\bm{\theta};(\bm{x}_i,\bm{y}_i))
    = \frac{1}{n} \sum_{i\in [n]} f_i(\bm{\theta}).
\end{align*}
In this paper, we focus on finding a stationary point $\bm{\theta}^\star \in \mathbb{R}^d$ such that $\nabla f(\bm{\theta}^\star) = \bm{0}$.

The loss functions $f_i$ ($i\in [n]$) satisfy the conditions in Assumption~\ref{assum:1}.

\begin{assumption}\label{assum:1}
Let $n \in \mathbb{N}$ be the number of training samples, and let $L_i > 0$ for all $i \in [n]$.

\begin{description}
  \item[(A1)] Each loss function $f_i \colon \mathbb{R}^d \to \mathbb{R}$ is differentiable and $L_i$-smooth. That is, for all $\bm{\theta}_1, \bm{\theta}_2 \in \mathbb{R}^d$,
  \[
    \|\nabla f_i(\bm{\theta}_1) - \nabla f_i(\bm{\theta}_2)\| \leq L_i \|\bm{\theta}_1 - \bm{\theta}_2\|.
  \]
  We also assume \( f_i^\star := \inf\{f_i(\bm{\theta}) : {\bm{\theta} \in \mathbb{R}^d}\} \in \mathbb{R} \).

  \item[(A2)] Let \(\xi\) be a random variable independent of \(\bm{\theta} \in \mathbb{R}^d\).  
  $\nabla f_{\xi} \colon \mathbb{R}^d \to \mathbb{R}^d$ is the stochastic gradient of $\nabla f$ that satisfies
  \begin{align*}
    &\textnormal{(i)} \quad \mathbb{E}_{\xi}[\nabla f_{\xi}(\bm{\theta})] = \nabla f(\bm{\theta}), \\
    &\textnormal{(ii)} \quad \mathbb{E}_{\xi}\left[\| \nabla f_{\xi}(\bm{\theta}) - \nabla f(\bm{\theta}) \|^2\right] \leq \sigma^2
  \end{align*}
  for some \(\sigma \geq 0\) and all \(\bm{\theta} \in \mathbb{R}^d\).

  \item[(A3)] Let $b \in \mathbb{N}$ such that $b \leq n$, and let $\bm{\xi} = (\xi_{1}, \xi_{2}, \cdots, \xi_{b})^\top$ comprise $b$ independent and identically distributed variables. The full gradient $\nabla f (\bm{\theta})$ is then estimated using the mini-batch gradient at $\bm{\theta}$:
  \begin{align*}
    \nabla f_{B}(\bm{\theta}) := \frac{1}{b} \sum_{i=1}^b \nabla f_{\xi_{i}}(\bm{\theta})
  \end{align*} 
where $\bm{\xi}$ is independent of $\bm{\theta} \in \mathbb{R}^d$.
\end{description}
\end{assumption}

\subsection{Mini-batch SGD}

At each iteration $t \in \mathbb{N}$, given the current parameter $\bm{\theta}_t \in \mathbb{R}^d$, mini-batch SGD selects $b_t$ loss functions $f_{\xi_{t,1}},\cdots,f_{\xi_{t,b_t}}$ randomly from $\{f_1,\cdots,f_n\}$, where $\bm{\xi}_t = (\xi_{t,1}, \cdots, \xi_{t,b_t})^\top$ is independent of $\bm{\theta}_t$ and $b_t$ is a batch size satisfying $b_t \leq n$.
The pseudo-code for the algorithm is shown as Algorithm \ref{algo:1}.

\begin{algorithm}
\caption{Mini-batch SGD algorithm}
\label{algo:1}
\begin{algorithmic}[1]
\REQUIRE
$\bm{\theta}_0 \in \mathbb{R}^d$ (initial point), 
$b_t > 0$ (batch size), 
$\eta_t > 0$ (learning rate), 
$T \geq 1$ (steps)
\ENSURE 
$(\bm{\theta}_t) \subset \mathbb{R}^d$
\FOR{$t=0,1,\ldots,T-1$}
\STATE{ 
$\nabla f_{B_t}(\bm{\theta}_t)
:=
\frac{1}{b_t} \sum_{i=1}^{b_t} \nabla f_{\xi_{t,i}}(\bm{\theta}_t)$}
\STATE{
$\bm{\theta}_{t+1} 
:= \bm{\theta}_t - \eta_t \nabla f_{B_t}(\bm{\theta}_t)$}
\ENDFOR
\end{algorithmic}
\end{algorithm}

The following lemma can be proved using Assumption \ref{assum:1} and the descent lemma \citep[Lemma 5.7]{beck2017}: for all $\bm{\theta}_1, \bm{\theta}_2 \in \mathbb{R}^d$,
\begin{align*}
f(\bm{\theta}_2) 
\leq f(\bm{\theta}_1) 
+ \langle \nabla f(\bm{\theta}_1), \bm{\theta}_2 - \bm{\theta}_1 \rangle
+ \frac{L}{2} \|\bm{\theta}_2 - \bm{\theta}_1\|^2,
\end{align*}
where Assumption \ref{assum:1} (A1) ensures that $f$ is $L$-smooth, with $L := \frac{1}{n} \sum_{i\in [n]} L_i$. The proof is given in \citet{umeda2025increasing}.

\begin{lemma}\label{lem:1}
Suppose Assumption \ref{assum:1} holds and consider the sequence $(\bm{\theta}_t)$ generated by Algorithm \ref{algo:1} with $\eta_t \in [\eta_{\min}, \eta_{\max}] \subset [0, \frac{2}{L})$ satisfying $\sum_{t=0}^{T-1} \eta_t \neq 0$, where 
$L := \frac{1}{n} \sum_{i\in [n]} L_i$ and $f^\star := \frac{1}{n} \sum_{i\in [n]} f_i^\star$.
Then, for all $T \in \mathbb{N}$,
\begin{align*}
&\min_{t\in [0:T-1]} \mathbb{E} \left[\|\nabla f(\bm{\theta}_t)\|^2 \right]
\\&\leq \frac{2(f(\bm{\theta}_0) - f^\star)}{2 - L \eta_{\max}}
\frac{1}{\sum_{t=0}^{T-1} \eta_t}
+ 
\frac{L \sigma^2}{2 - L \eta_{\max}} 
\frac{\sum_{t=0}^{T-1} \eta_t^2 b_t^{-1}}{\sum_{t=0}^{T-1} \eta_t},
\end{align*}
where $\mathbb{E}$ denotes the total expectation, defined by $\mathbb{E} := \mathbb{E}_{\bm{\xi}_0}\mathbb{E}_{\bm{\xi}_1} \cdots \mathbb{E}_{\bm{\xi}_t}$.
\end{lemma}

\begin{table}[htbp]
\centering
\begin{tabular}{lc}
\toprule
Scheduling strategy
&
$\min_{t} \mathbb{E}[\|\nabla f(\bm{\theta}_t)\|]$\\
\midrule
\textbf{(i)~$b_t$:\ Increase;\ $\eta_t$\ : Constant}
&
$\displaystyle{O \left( \frac{1}{\sqrt{T}} \right)}, \displaystyle{O \left( \frac{1}{\sqrt{M}} \right)}$ \\
\midrule
\textbf{(ii)~$b_t$:\ Increase;\ $\eta_t$\ : Increase}
&
$\displaystyle{O \left( \frac{1}{\gamma^{\frac{M}{2}}} \right)}$ \\
\bottomrule              
\end{tabular}
\caption{
Theoretical upper bounds of $\min_t \mathbb{E}[\|\nabla f(\bm{\theta}_t)\|]$ under two scheduling strategies~\citep{umeda2025increasing}.
Here, $T$ denotes the total number of optimization steps, 
$M$ the number of times the batch size is increased during training, 
and $\gamma>1$ is the learning rate growth factor defined in \eqref{eq:exp-scheduler}.
}
\label{tab:scheduler_bounds}
\end{table}

Building on Lemma~\ref{lem:1}, \citet{umeda2025increasing} conducted a convergence analysis of various batch-size and learning-rate scheduling strategies.
Their results, summarized in Table~\ref{tab:scheduler_bounds}, theoretically demonstrate that increasing the batch size improves the convergence rate, offering a clear advantage over fixed-batch training.
Moreover, the convergence rates in Table~\ref{tab:scheduler_bounds} indicate that jointly increasing both the batch size and learning rate yields even faster convergence.

\subsection{SFO Complexity}

First-order optimizers, such as SGD and its variants, use stochastic gradients estimated from mini-batches of training data. A fundamental metric in this context is \emph{SFO complexity}, defined as the total number of gradient computations during training. For batch size $b$ and total number of iterations $T$, SFO complexity is given by 
\[
    N := T b.
\]
SFO complexity quantifies the total computational effort required to reach an $\epsilon$–approximate stationary point, typically defined by
\begin{align*}
    \min_{t \in [0:T-1]} \mathbb{E}[\|\nabla f(\bm{\theta}_t)\|] \leq \epsilon.
\end{align*}
Under Assumption~\ref{assum:1}, existing analyses have established upper bounds of the form
\begin{align} \label{eq:grad-bound}
    \min_{t \in [0:T-1]} \mathbb{E}[\|\nabla f(\bm{\theta}_t)\|^2] \leq \frac{C_1(\eta)}{T} + \frac{C_2(\eta)}{b}, 
\end{align}
where 
\[
C_1(\eta) := \frac{2(f(\bm{\theta}_0)-f^\star)}{(2-L\eta)\eta}, 
\quad 
C_2(\eta) :=\frac{L\sigma^2\eta}{2-L\eta}
\]
depend on the constant learning rate $\eta$, the Lipschitz constant $L$, and the gradient noise variance $\sigma^2$ \citep{Imaizumi19062024}.  

From \eqref{eq:grad-bound}, reaching an $\epsilon$–approximate stationary point requires
$\frac{C_1(\eta)}{T} + \frac{C_2(\eta)}{b} \le \epsilon^2$.
To obtain the minimal number of iterations, we consider the case in which the inequality holds with equality, i.e.,
\[
\frac{C_1(\eta)}{T} + \frac{C_2(\eta)}{b} = \epsilon^2.
\]
Solving this inequality for the number of iterations $T$ yields
\begin{align}\label{eq:T-of-b}
    T(b, \eta) = \frac{C_1(\eta) b}{\epsilon^2 b - C_2(\eta)}, 
    \quad \left(b > \frac{C_2(\eta)}{\epsilon^2}\right).
\end{align}
Substituting $T(b, \eta)$ into the definition of SFO complexity, $N(b, \eta) = T(b, \eta) \cdot b$, directly gives
\begin{align}\label{eq:sfo-explicit}
    N(b, \eta) = \frac{C_1(\eta)b^2}{\epsilon^2 b - C_2(\eta)}, 
    \quad \left(b > \frac{C_2(\eta)}{\epsilon^2}\right).
\end{align}

Recent work has shown empirically that there exists a \emph{critical batch size} $b^\star$ that balances computational efficiency and optimization dynamics \citep{mccandlish2018empiricalmodellargebatchtraining, pmlr-v80-ma18a, shallue2019, zhang2025how}.  
\citet{Imaizumi19062024} further formalized this phenomenon by showing that the number of iterations $T(b, \eta)$ required to reach an $\epsilon$–approximate stationary point is a decreasing and convex function of batch size $b$.  
Consequently, SFO complexity $N(b, \eta) = T(b, \eta) \cdot b$ is itself a convex function in $b$ and has a unique minimizer at which the derivative vanishes; that is, $N'(b^\star) = 0$.  
The critical batch size that minimizes \eqref{eq:sfo-explicit} is then obtained as
\begin{align} \label{eq:critical-batch-size}
    b^\star = \frac{2 C_2(\eta)}{\epsilon^2}.
\end{align}
This result provides a theoretical justification for the empirically observed \emph{critical batch size} $b^\star$. Increasing the batch size beyond this point yields diminishing returns in terms of training efficiency as the benefit of variance reduction is offset by the increased computational cost.

It is also important to note that, for a fixed number of \emph{epochs}, total SFO complexity does not depend on the batch size.
Indeed, if the dataset size is $n$ and the batch size is $b$, the number of iterations in one epoch is $T_e = \lceil {n}/{b}\rceil$,
and each update incurs a cost proportional to $b$.
Hence, total SFO complexity per epoch $N_{e}$ is given by
\[
N_{e} = T_e \, b = \left\lceil \frac{n}{b} \right\rceil b \approx n,
\]
which corresponds to the total number of samples processed in one pass through the dataset.
As a result, under epoch budget $E$, total SFO complexity is given by
\[
N = Tb = E\cdot T_eb =  E \cdot \left\lceil \frac{n}{b} \right\rceil b \approx E \cdot n,
\]
showing that it scales linearly with the number of epochs $E$ but is nearly independent of batch size $b$.

Therefore, when comparing different batch size schedules,
the key indicator of training efficiency is how much the gradient norm can be reduced for a fixed number of epochs.
In other words, for a given epoch count, the schedule that achieves the smallest value of $\min_t \|\nabla f(\bm{\theta}_t)\|$  utilizes a fixed SFO complexity most effectively.

\section{Optimal Growth Schedules for Batch Size and Learning Rate}

\subsection{Design of Batch Size and Learning Rate Schedules}

For each stage $m \in [0, M)$, we fix batch size $b_m$ and learning rate $\eta_m$ and partition the training process into $M$ stages.
Let $T_m$ denote the cumulative iteration count up to the end of stage $m$ 
(with $T_{-1}=0$), and let $\Delta T_m := T_m - T_{m-1}$ denote the stage length.
In each stage, batch size $b_m$ and learning rate $\eta_m$ remain constant.
Then, for each stage $m$, the standard nonconvex convergence bound~\eqref{eq:grad-bound} yields
\begin{align*}
    \min_{t \in [T_{m-1}, T_m)}\mathbb{E}\|\nabla f(\bm{\theta}_t)\|^2 \leq\frac{C_1(\eta_m)}{\Delta T_m} + \frac{C_2(\eta_m)}{b_m},
\end{align*}
where the constants are given by
\[
    C_1(\eta_m) := \frac{2\left(f(\bm{\theta}_{T_{m-1}})-f^\star\right)}{(2-L\eta_m)\eta_m},
    C_2(\eta_m) := \frac{L\sigma^2\eta_m}{2-L\eta_m}.
\]
If we target an accuracy level $\epsilon$ for each stage, the number of iterations $\Delta T_m$ required in stage $m$ satisfies
\[
    \Delta T_m = \frac{C_1(\eta_m) b_m}{\epsilon^2 b_m - C_2(\eta_m)}.
\]
The \emph{per-stage SFO complexity} is obtained by multiplying $\Delta T_m$ by batch size $b_m$:
\begin{align*}
    N(b_m,\eta_m) :&= b_m\,\Delta T_m \\
    &= \frac{C_1(\eta_m)\,b_m^2}{\epsilon^2 b_m - C_2(\eta_m)}, \left(b_m > \frac{C_2(\eta_m)}{\epsilon^2}\right).
\end{align*}
Finally, total SFO complexity is obtained by summing over all stages $m = 0, 1, \dots, {M-1}$:
\begin{align*}
    N = \sum_{m=0}^{M-1} N(b_m,\eta_m).
\end{align*}
Moreover, each stage admits a \emph{critical batch size} that minimizes 
$N(b_m,\eta_m)$:
\begin{align}\label{eq:critical-bm}
    b_m^\star = \frac{2\,C_2(\eta_m)}{\epsilon^2}.
\end{align}
Thus, the optimal batch size schedule should track the increase in the per-stage critical batch size $b_m^\star$, which depends on both the current learning rate $\eta_m$ and the target accuracy $\epsilon$.

\subsubsection{Increasing Batch Size with Constant Learning Rate}
Table~\ref{tab:scheduler_bounds} (i) shows that the upper bound of $\min_t \mathbb{E}[\|\nabla f(\bm{\theta}_t)\|]$ decays at a rate of $O(1/\sqrt{T})$ when the batch size is increased and the learning rate is kept constant ($\eta_m = \eta$). Since the total number of steps across $M$ batch size increases satisfies $T_M = \sum_{m=0}^{M-1} \Delta T_m \geq M$, the convergence rate $O({1}/{\sqrt{T}})$ can be equivalently expressed as $O({1}/{\sqrt{M}})$.
Therefore, since \(\epsilon^2\) decreases as \(O(1/M)\), it follows from \eqref{eq:critical-bm} that the critical batch size \(b_m^\star\) scales as \(O(M)\).  
In other words, the critical batch size increases linearly with \(M\).
Hence, adopting a \emph{linear growth} batch size schedule yields\\

\textbf{[Linear Growth BS]}
\begin{equation} \label{eq:linear-scheduler}
    b_m = b_0 + m \cdot \Delta b,
\end{equation}\\
where $\Delta b \in \{n \in \mathbb{Z} \mid n \geq 0\}$. This schedule matches the scaling behavior of the critical batch size.

\subsubsection{Exponential Growth of both Batch Size and Learning Rate}

Table~\ref{tab:scheduler_bounds} (ii) shows that the upper bound of
$\min_t \mathbb{E}[\|\nabla f(\bm{\theta}_t)\|]$ decays at a rate of $O(\gamma^{-M/2})$
when both the batch size and learning rate are increased exponentially.
Hence, adopting an \emph{exponential growth} schedule for both the batch size and learning rate yields\\

\textbf{[Exponential Growth BS and LR]}
\begin{equation}
\label{eq:exp-scheduler}
b_m = b_0 \cdot \delta^m, \quad
\eta_m = \eta_0\cdot\gamma^m,
\end{equation}\\
where $\delta, \gamma > 1$ and $\gamma^2 < \delta$. In this setting, for $\eta_m \leq 1/L$, the term $C_2(\eta_m)$ satisfies
\begin{equation}
C_2(\eta_m) = \frac{L\sigma^2 \eta_m}{2 - L\eta_m} \leq L\sigma^2\eta_m,
\end{equation}
which increases as $O(\gamma^M)$.
Meanwhile, the target accuracy $\epsilon^2$ decays as $O(\gamma^{-M})$.
Substituting these scaling behaviors into the critical batch size expression \eqref{eq:critical-bm} shows that the critical batch size $b_m^\star$ increases as $O(\gamma^{2M})$.

Meanwhile, the scheduled batch size $b_m$ increases as $O(\delta^M)$.
To match the growth of the critical batch size, it is necessary that $\gamma^2 \approx \delta$.
Equivalently, setting $\gamma \approx \sqrt{\delta} \quad (\text{with } \gamma < \sqrt{\delta})$ ensures that $b_m$ increases at nearly the same rate as $b_m^\star$.
If $\gamma$ is set smaller than $\sqrt{\delta}$, the scheduled batch size $b_m$ increases faster than necessary, leading to an unnecessary increase in SFO complexity and a corresponding reduction in training efficiency.

Next, consider the term $C_1(\eta_m)$, given by
\begin{equation}
C_1(\eta_m) =
\frac{2(f(\bm{\theta}_{T_{m-1}})-f^\star)}{(2 - L\eta_m)\eta_m}.
\end{equation}
This function is convex in $\eta_m$ and reaches its minimum at $\eta_m = 1/L$.
Thus, $C_1(\eta_m)$ decreases monotonically for $\eta_m \leq 1/L$ but increases once $\eta_m > 1/L$.
Therefore, exceeding $\eta_m > 1/L$ results in a larger $C_1(\eta_m)$, thereby increasing SFO complexity.

From these observations, it is preferable to maintain $\eta_m \leq 1/L$.
Although the convergence of SGD is theoretically guaranteed for the broader range $\eta_m < 2/L$, minimizing SFO complexity requires progressively increasing $\eta_m$ while maintaining $\eta_m \leq 1/L$.

To translate these theoretical insights into a practical training procedure, we use a mini-batch SGD framework that updates the batch size and learning rate at each stage in accordance with the derived schedules.
Specifically, we designed an algorithm that tracks the current stage $m$, updates $b_m$ and $\eta_m$ following either the linear growth schedule \eqref{eq:linear-scheduler} or the exponential growth schedule \eqref{eq:exp-scheduler}, and iterates for a prescribed number of epochs per stage.
The full procedure for the exponentially increasing schedule is summarized in Algorithm~\ref{algo:2} below.

\begin{algorithm}[htbp]
\caption{Mini-batch SGD Algorithm with Exponential Growth BS and LR Schedule}
\label{algo:2}
\begin{algorithmic}[1]
\REQUIRE 
$\bm{\theta}_0 \in \mathbb{R}^d$ (initial parameters), 
$b_0 > 0$ (initial batch size), 
$\eta_0 > 0$ (initial learning rate), 
$M \ge 1$ (number of stages), 
$\delta, \gamma > 1$ (growth factors),
$n \geq 1$ (number of training samples),
$E \geq 1$ (epochs per stage)
\ENSURE
$(\bm{\theta}_t) \subset \mathbb{R}^d$
\STATE $t \gets -1$
\FOR{$m = 0,1,\dots,M-1$} 
    \STATE $b_m \gets b_0 \cdot \delta^m$
    \STATE $\eta_m \gets \eta_0 \cdot \gamma^m$
    \STATE $\Delta T_m = \left\lceil{n}/{b_m}\right\rceil \cdot E$
    \FOR{$i = 1,\dots,\Delta T_m$}
        \STATE $t \gets t + 1$
        \STATE $\nabla f_{B_t}(\bm{\theta}_t) := \frac{1}{b_m}\sum_{j = 1}^{b_m} \nabla f_{\xi_{t, j}}(\bm{\theta}_t)$
        \STATE $\bm{\theta}_{t+1} := \bm{\theta}_t - \eta_m \nabla f_{B_t}(\bm{\theta}_t)$
    \ENDFOR
\ENDFOR
\end{algorithmic}
\end{algorithm}
\section{Evaluation}

To evaluate the effectiveness of our scheduling strategies, we performed experiments using Algorithms~\ref{algo:1} and~\ref{algo:2} to train ResNet-18 on the CIFAR-100 dataset.
All experiments were conducted on a system equipped with an NVIDIA A100 40-GB GPU and an AMD EPYC 7742 2.25-GHz CPU.
The software stack comprised Python 3.10.12, PyTorch 2.1.0, and CUDA 12.2.

We set the total number of epochs $E = 200$ and the initial learning rate $\eta_0 = 0.1$.

\subsection{Effectiveness of Different Batch Size Growth Schedules with Fixed Learning Rate}

\begin{figure}[htbp]
  \centering
    \includegraphics[page=1,width=\linewidth]{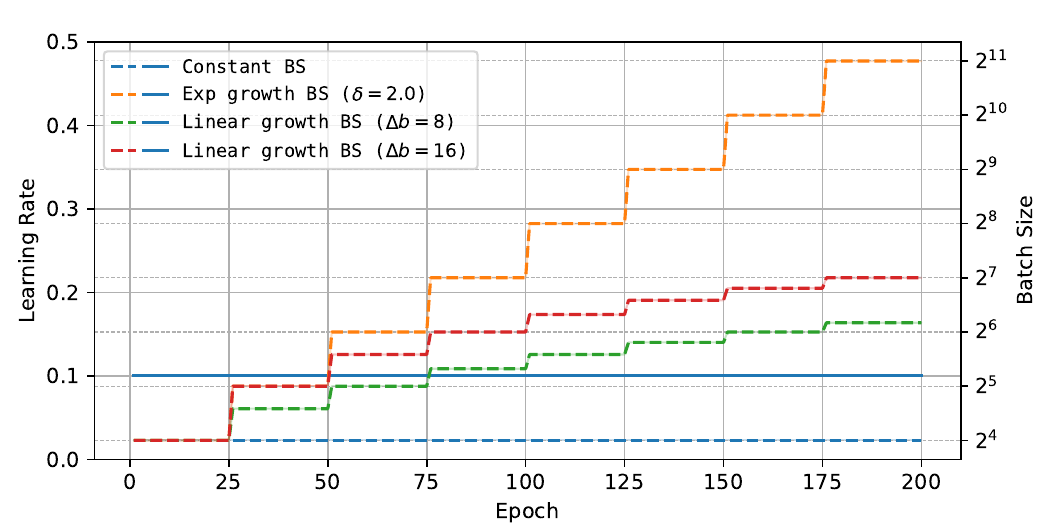}
    \caption*{(a) Learning Rate and Batch Size Schedule}
    \includegraphics[page=2,width=\linewidth]{figures/cifar100_comp_linear.pdf}
    \caption*{(b) Full Gradient Norm of Empirical Loss for Training}
    \includegraphics[page=3,width=\linewidth]{figures/cifar100_comp_linear.pdf}
    \caption*{(c) Empirical Loss Value for Training}
    \includegraphics[page=4,width=\linewidth]{figures/cifar100_comp_linear.pdf}
    \caption*{(d) Accuracy Score for Test}
  \caption{Comparison of performance with fixed learning rate ($\eta = 0.1$) and four batch size schedules: (i) constant ($b=16$), (ii) exponential growth ($\delta=2.0$), (iii) linear growth ($\Delta b=8$), (iv) linear growth ($\Delta b=16$).}
  \label{fig:compare-steps-linear}
\end{figure}

We first consider the case in which the learning rate is kept constant ($\eta = 0.1$) and the batch size is either kept constant, increased linearly in accordance with \eqref{eq:linear-scheduler}, or increased exponentially in accordance with \eqref{eq:exp-scheduler}.
Figure~\ref{fig:compare-steps-linear} (a) plots the batch size and learning rate schedules for each alternative. The solid line indicates the mean value, and the shaded area indicates the range between maximum and minimum across three runs. 

The results plotted in Figures~\ref{fig:compare-steps-linear} (b)--(d) indicate that increasing the batch size improves convergence with respect to SFO complexity compared with keeping it fixed.
In particular, the linear growth schedule \eqref{eq:linear-scheduler} closely tracks the critical batch size at each stage, resulting in a steady reduction of the gradient norm throughout training.

In contrast, the exponential growth schedule \eqref{eq:exp-scheduler} increases the batch size too aggressively, causing it to exceed the critical batch size prematurely. This results in a slowdown in the reduction of the gradient norm during later stages and higher SFO complexity.

When the batch size is fixed, it fails to follow the increasing critical batch size, resulting in slower convergence.

\subsection{Effectiveness of Different Learning Rate Growth Schedules with Fixed Exponential Batch Size Growth}

\begin{figure}[htbp]
  \centering
    \includegraphics[page=1,width=\linewidth]{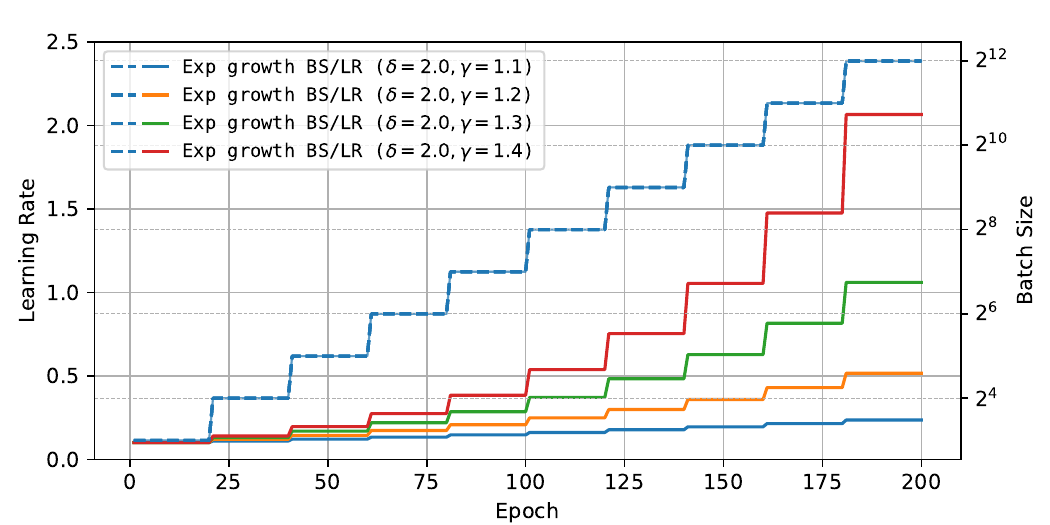}
    \caption*{(a) Learning Rate and Batch Size Schedule}
    \includegraphics[page=2,width=\linewidth]{figures/cifar100_comp_lr.pdf}
    \caption*{(b) Full Gradient Norm of Empirical Loss for Training}
    \includegraphics[page=3,width=\linewidth]{figures/cifar100_comp_lr.pdf}
    \caption*{(c) Empirical Loss Value for Training}
    \includegraphics[page=4,width=\linewidth]{figures/cifar100_comp_lr.pdf}
    \caption*{(d) Accuracy Score for Test}
  \caption{Comparison of performance when batch size is exponentially increased with a fixed growth factor ($\delta = 2.0$) and learning rate is exponentially increased with various growth factors ($\gamma = 1.1, 1.2, 1.3, 1.4$).}
  \label{fig:compare-steps-lr}
\end{figure}

Next, we consider the case in which the batch size is exponentially increased with a fixed growth factor ($\delta = 2.0$), as defined in \eqref{eq:exp-scheduler}, while the learning rate schedule follows \eqref{eq:exp-scheduler} with $\gamma = 1.1, 1.2, 1.3,$ or $1.4$.
Figure~\ref{fig:compare-steps-lr} (a) plots the batch size and learning rate schedules for each alternative. The solid line indicates the mean value, and the shaded area indicates the range between maximum and minimum across three runs.

The results plotted in Figures~\ref{fig:compare-steps-lr} (b)--(d) indicate that a larger learning rate growth factor $\gamma$ (with $\gamma < \sqrt{\delta}$) results in a smaller gradient norm, consistent with the theoretical complexity of $O(\gamma^{M/2})$. 

Among the tested settings, $\gamma = 1.4$ yields the best convergence as it approximately satisfies $\gamma \approx \sqrt{\delta}$, thereby synchronizing the growth in the learning rate and batch size with the critical batch size at each stage.

\subsection{Effectiveness of Different Batch Size Growth Schedules with Fixed Learning Rate Growth}

\begin{figure}[htbp]
  \centering
    \includegraphics[page=1,width=\linewidth]{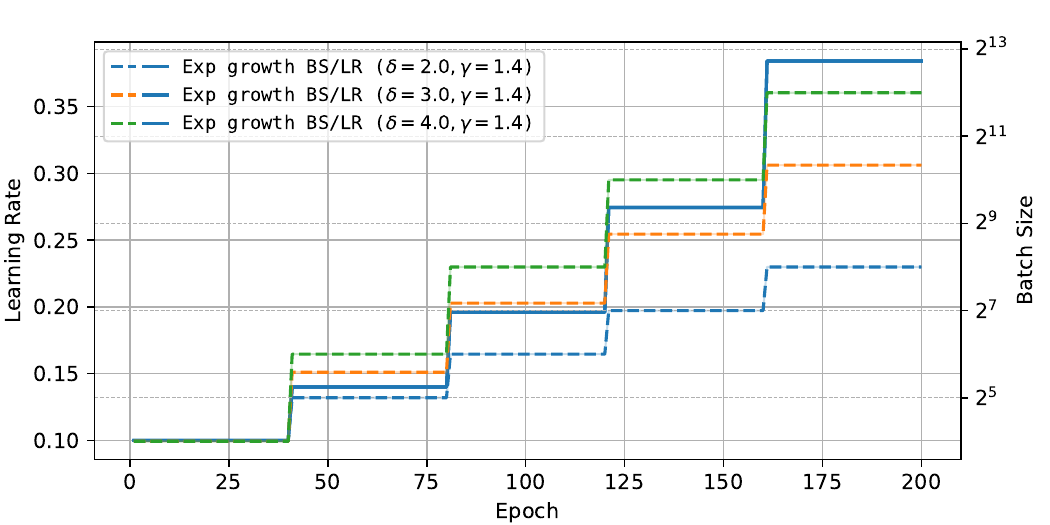}
    \caption*{(a) Learning Rate and Batch Size Scheduler}
    \includegraphics[page=2,width=\linewidth]{figures/cifar100_comp_bs.pdf}
    \caption*{(b) Full Gradient Norm of Empirical Loss for Training}
    \includegraphics[page=3,width=\linewidth]{figures/cifar100_comp_bs.pdf}
    \caption*{(c) Empirical Loss Value for Training}
    \includegraphics[page=4,width=\linewidth]{figures/cifar100_comp_bs.pdf}
    \caption*{(d) Accuracy Score for Test}
  \caption{Comparison of performance when learning rate is exponentially increased with a fixed growth factor ($\gamma = 1.4$) and batch size is exponentially increased with various growth factors ($\delta = 2.0, 3.0, 4.0$).}
  \label{fig:compare-steps-bs}
\end{figure}

Finally, we consider the case in which the learning rate is exponentially increased with a fixed growth factor ($\gamma = 1.4$), as defined in \eqref{eq:exp-scheduler}, while the batch size follows \eqref{eq:exp-scheduler} with $\delta = 2.0, 3.0,$ and $4.0$.
Figure~\ref{fig:compare-steps-bs} (a) plots the batch size and learning rate schedules for each alternative. The solid line indicates the mean value, and the shaded area indicates the range between maximum and minimum across three runs.

The results plotted in Figures~\ref{fig:compare-steps-bs} (b)--(d) reveal that all settings exhibit the same theoretical convergence rate, $O(\gamma^{M/2})$, and that the actual gradient norm is lowest when batch size growth factor $\delta$ is smaller.
This is because, under the fixed $\gamma = 1.4$ setting, the $\delta = 2.0$ setting satisfies $\gamma \approx \sqrt{\delta}$, aligning the scheduled batch size with the critical batch size at each stage. This results in a more efficient reduction in SFO complexity compared with using larger values of $\delta$.

\section{Conclusion}

In this work, we investigated how jointly increasing the batch size and learning rate can enhance the training efficiency of mini-batch stochastic gradient descent (SGD) while preserving convergence guarantees.
By analyzing the problem through the lens of stochastic first-order oracle complexity, we derived theoretical conditions for optimal growth schedules and identified the \emph{critical batch size} that minimizes computational cost at each stage.
Our analysis showed that, for exponential schedules, the optimal relationship between growth factors is approximately $\gamma^2 \approx \delta$, ensuring that the scheduled batch size grows in sync with the per-stage critical batch size.

We validated these insights through extensive experiments on ResNet‑18 with the CIFAR‑100 dataset, confirming that carefully designed schedules significantly improve convergence efficiency.
In particular, linear batch size growth closely tracks the increasing critical batch size under a constant learning rate, while exponential schedules achieve even faster convergence when the learning rate and batch size are coupled in accordance with the theoretical relation.
These results provide actionable guidelines for practitioners, demonstrating how to balance batch size and learning rate dynamics to fully leverage GPU parallelism without incurring unnecessary computational overhead or compromising generalization.

Beyond improving the efficiency of mini-batch SGD, our findings offer a principled foundation for designing scalable training strategies for large-scale deep learning.
Future work will extend this analysis to adaptive optimizers such as Adam, investigate schedule design under non-stationary noise conditions and heavy-tailed gradient distributions, and explore automatic schedule tuning based on online estimation of the critical batch size.

\bibliography{aaai2026}

\end{document}